\pdfoutput=1
\documentclass[11pt]{article}

\usepackage[]{ACL2023}

\usepackage{times}
\usepackage{latexsym}

\usepackage[T1]{fontenc}
\usepackage[utf8]{inputenc}

\usepackage{microtype}

\usepackage{inconsolata}

\usepackage{graphicx}
\usepackage[english]{babel}

\newcommand{\exrole}[1]{{\tt {#1}}}

\title{Extending an Event-type Ontology: Adding Verbs and Classes\\
Using Fine-tuned LLMs Suggestions}
\author{Jana Straková \and Eva Fučíková \and Jan Hajič \and Zdeňka Urešová\\
  Institute of Formal and Applied Linguistics, Faculty of Mathematics and Physics \\
  Charles University, Prague, Czech Republic\\
  \texttt{\{strakova,fucikova,hajic,uresova\}@ufal.mff.cuni.cz}}

\usepackage[absolute]{textpos}
\begin{document}
\begin{textblock}{16}(0,0.1)\centerline{This paper was published at \textbf{LAW-XVII @ ACL 2023}. Please cite the published version: \small{\url{https://aclanthology.org/2023.law-1.9/}}}\end{textblock}
\maketitle
\begin{abstract}
In this project, we have investigated the use of advanced machine learning methods, specifically fine-tuned large language models, for pre-annotating data for a lexical extension task, namely adding descriptive words (verbs) to an existing (but incomplete, as of yet) ontology of event types. Several research questions have been focused on, from the investigation of a possible heuristics to provide at least hints to annotators which verbs to include and which are outside the current version of the ontology, to the possible use of the automatic scores to help the annotators to be more efficient in finding a threshold for identifying verbs that cannot be assigned to any existing class and therefore they are to be used as seeds for a new class. We have also carefully examined the correlation of the automatic scores with the human annotation. While the correlation turned out to be strong, its influence on the annotation proper is modest due to its near linearity, even though the mere fact of such pre-annotation leads to relatively short annotation times.

\end{abstract}

\section{Introduction}

Annotation of highly-dimensional, voluminous data is expensive, time-consuming and in addition, in case of deep-niche domains, depending on expertly trained specialists, such as linguists or medical experts. Therefore it may be advantageous to organize, prioritize and provide suggestions to guide further annotation efforts efficiently. Especially in a situation with a rich, constantly growing set of classes, such as it is the case with ontologies. 

% JS: Tahle věta se nám nějak opakuje a shoduje s následující větou, tak jsem ten druhý výskyt dala pryč.
% A dokonce to byla připomínka Reviewera #1:
% Lines 035-038 and 039-41: duplicates
Specifically, given an already partially labeled set of examples with yet unfinished set of classes, classifier based on large language models (LLMs) can be leveraged to navigate the landscape of possible annotations. 

Our showcase is an event-type ontology, the SynSemClass 4.0 \citep{uresova-etal-2022-making}, populated with synonymous verbs denoting events or states. The set of events is currently dynamically evolving and encompasses classes in English, Czech, German and Spanish, so far limited to verbs.

As any ontological resource is never complete, we have investigated various methods to facilitate efficient extension of such ontologies in two ways: adding classes for greater coverage on new texts, and adding verbs to existing classes to allow for more accurate human understanding of the classes in the ontology for a particular form of the given class expression.

\noindent We suggest to achieve these by

\begin{enumerate}
    \item examining examples with consistently low class affiliation scores across a large corpus as potential candidates for new classes;
    \item on the other side of the spectrum, examining high-certainty decisions of a supervised classifier to locate highly-affiliated lemmas to a particular class, corresponding to ``low-hanging fruit'' for a quick manual review and confirmation of the inclusion of the lemma into the suggested class.
\end{enumerate}

In all cases, classifier prediction serves as guidance and the annotators are briefed to consider the suggestions as election votes. The final decision is always the annotator's, who can accept or dismiss the suggestions. 

The organization of this paper is as follows: Sect.~\ref{sec:ontology} introduces the SynSemClass v4 ontology and the current state of annotations. Sect.~\ref{sec:generating} describes the fine-tuned LLM classifier used to generate the annotation suggestions. Sect.~\ref{sec:manual-annotation} describes the manual annotations post-processing. Results are presented in Sect.~\ref{sec:results} and discussed in Section~\ref{sec:discussion}. Finally, we conclude in Sect.~\ref{sec:conclusions}.

We release the source code at \url{https://github.com/strakova/synsemclass_ml}.

\section{The Ontology}
\label{sec:ontology}

In our experiments, we have used the Czech part of the SynSemClass 4.0\footnote{\url{https://lindat.mff.cuni.cz/repository/xmlui/handle/11234/1-4746}} \citep{uresova-etal-2022-making} in which
% JS in which vs where: https://prowritingaid.com/in-which-vs-where (where jen pro místa myslím)
contextually-based synonymous verbs in various languages are classified into multilingual synonym classes according to the semantic and syntactic properties they display. There is no specific model or lexicographic theory behind building the database. However, from the linguistic point of view, the notion of synonymy used is based on the ``loose'' definition of synonymy by Lyons and Jackson \cite{Lyons1968,Jackson1988}, or alternatively and very closely, on both ``near-synonyms'' and ``partial synonyms'' as defined by Lyons \cite{Lyons1995,cruse2000} or ``plesionyms'' as defined by Cruse \cite{cruse}.\footnote{The ``loose'' definition of synonymy covers synonyms that fulfil some of the conditions stipulated for synonymy in the strictest sense but not all and does not work with the ``absolute'' synonymy covering the total identity of meaning.
%ZU: %ZU: 092-099: please briefly explain what is at the core of these different definitions of synonyms. doplňuji:
%JS: Přesunula jsem komentář tak, aby footnote následovala hned za tečkou, jinak vzniká nadbytečná mezera.
The ``partial synonymy'' is defined \cite{Lyons1995} as a relationship holding between two lexemes that satisfy the criterion of identity of meaning, but do not meet all the conditions of absolute synonymy. The ``near-synonymy'' \cite{Lyons1995} and ``plesionyms'' \cite{cruse} is defined as ``expressions that are more or less similar, but not identical, in meaning''.}

From the ontological point of view, the classes are meant to reflect different event types (concepts) and collect various information about the possible forms of expression of the event type in language.

\begin{figure}[!htbp]
\begin{center}
\includegraphics[width=0.48\textwidth]{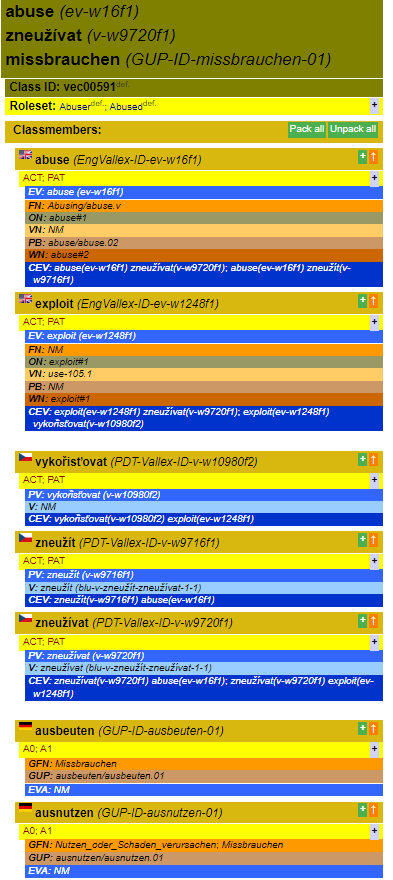}
\caption{SynSemClass example entry as presented on its public access website, Class ID: vec00591 (simplified)}
\label{slovnik2}
\end{center}
\end{figure}

The following main basic features are distinguished in SynSemClass (Fig.~\ref{slovnik2}) \cite{uresova-etal-2022-making}:
\begin{itemize}
    \item The \textbf{name of each multilingual class} stands for a single concept (e.g., of \textit{accelerating})\footnote{This is different from the commonly used term of ``semantic classes of verbs'' as represented, for example, in VerbNet, where the class is defined much more broadly -- such as for all verbs of movement.} and corresponds to the verb that represents the prototypical sense in each of the languages included: class member (CM) \textit{abuse} for English, \textit{zneužívat} for Czech, and \textit{missbrauchen} for German. So far, SynSemClass focuses on verbal synonyms since they carry the key syntactic-semantic information for language understanding.\footnote{As described in detail in \cite{Meaningand2019,LREC18czengclasspaper,TLT2018,CROSSLINGUAL2018}.}
 
  \item Each class is also provided with a brief language-dependent general \textbf{class definition}, which characterizes the meaning, or concept of the class, i.e. the meaning of all synonymous verbs contained in it. The class is viewed as a substitute for an ontology unit representing a single concept, similar to the treatment of WordNet synsets in \cite{wordnetRDF}. 

  \item For each class, SynSemClass also provides a fixed set (called \textbf{``Roleset''} (RS)) of defined ``situational participants'' (called ``semantic roles'' (SR)) that are common for all the members (the individual verb senses) of a particular class. The RS is mapped to the valency frame of the individual synonymous verbs securing for each synonymous verb to be characterized both meaning-wise (SR) and structurally (valency arguments). For example, the class {\tt vec00591} \textit{abuse}, as concept of ``abusing'', has two semantic roles, \exrole{Abuser} and \exrole{Abused} (Fig.~\ref{slovnik2}). Every role in SynSemClass is provided with a brief language-dependent general \textbf{role definition} as well as every class. While the SRs resemble FrameNet's ``Frame Elements'' (and sometimes borrow their names from there), it should be pointed out that there is one fundamental difference: the SRs used in SynSemClass aim at being defined across the ontology, and not per class (as they would be if we follow the ``per frame'' approach used in FrameNet). 

  \item Each individual language-dependent synonymous verb included in a given class is called \textbf{Class Member} and for each new CM to be added, it must be possible, in the prototypical case, to create a mapping between its syntactic arguments and the roles in that class' RoleSet; see the example in our web-based lexicon (Fig.~\ref{slovnik2}).\footnote{The public web version is available at \url{https://lindat.cz/services/SynSemClass40/SynSemClass40.html}} Each CM of one class is denoted by a verb lemma and the valency frame ID which, roughly speaking, represents the particular verb sense.
%ZU: 144-148: it's confusing that the SRs of the example, Abuser and Abused, seem nevertheless also class-specific and not cross-ontology.

%\begin{figure}[htpb]
%\begin{center}
%\includegraphics[width=0.5\textwidth]{slovnik.jpg}
%\caption{SynSemClass entry for the Class ID:vec00337 (simplified)}
%\label{slovnik}
%\end{center}
%\end{figure}
%ZU L164: introduce abbreviation SSC done
\item Each CM is further linked to one, or more existing online lexical resources for each language to support, e.g., comparative studies, or any other possible research in the community. In SynSemClass (SSC), there exist \textbf{links} to e.g., Vallex\footnote{\url{https://hdl.handle.net/11234/1-3524}} for Czech, FrameNet\footnote{\url{https://framenet.icsi.berkeley.edu/fndrupal/}} and VerbNet\footnote{\url{https://uvi.colorado.edu/ and http://verbs.colorado.edu/verbnet/index.html}} for English, E-VALBU\footnote{\url{https://grammis.ids-mannheim.de/verbvalenz}} for German, AnCora\footnote{\url{http://clic.ub.edu/corpus/es/ancoraverb\_es}} for Spanish. Each Class Member is exemplified by instances of real texts (and their translations to English) extracted from translated or parallel corpora. Specifically, data is extracted from the Prague Czech-English Dependency Corpus (PCEDT)\footnote{\url{https://ufal.mff.cuni.cz/pcedt2.0/en/index.html}}  for Czech-English, from the Paracrawl  corpus\footnote{\url{https://opus.nlpl.eu/ParaCrawl.php}} for German-English and from the XSRL dataset\footnote{\url{https://catalog.ldc.upenn.edu/LDC2021T09}} for Spanish-English. 
\end{itemize}

%ZU:L175: what does "885 active" mean? Is this related to Fn.14?done
SynSemClass 4.0 includes 1200 classes (885 active after merging or deleting) with 8169 Class Members. All classes are annotated in Czech and English, 60 of them have also German annotation. Spanish is not included in the web version but is under construction \cite{Spanish2023}. 
%About 130 classes are currently annotated with Spanish synonyms \citep{Spanish2023}. For the next version (5.0), it is planned to add 400 new classes. 

\section{Generating Annotation Suggestions with Fine-tuned LLM Classifier}
\label{sec:generating}

\subsection{Data}

The Czech part of the SynSemClass ontology\footnote{SynSemClass 4.0 with additions annotated since the last v4.0 release.} yielded 12045 example sentences with 3313 unique verbs (lemmas) manually annotated in 965 classes.\footnote{We considered only active (not merged, not deleted) classes in the current state of SynSemClass (SSC) annotated since v4.0 release, and naturally, only those classes which are represented with at least one example sentence (to be used as LLM input).} We have split the data randomly in proportion 80/10/10 in a stratified train/dev/test split,\footnote{Stratified means forcing the distribution of the target variable, in our case the classes, to be equal among the train/dev/test split.} resulting in 9635/1205/1205 train/dev/test examples.

 Our input is a list of 3389 completely new, unseen verbs (lemmas) and our motivation is to differentiate:
 
\begin{itemize}
     \item verbs consistently poorly classified as class members of any of the existing classes, i.e., possible candidates for establishing new classes, 
     \item verbs highly affiliate to some of the existing classes, i.e., possible candidates for adding them as one of the verbs characterizing an existing class.
\end{itemize}
  
To obtain the classification score for each lemma-class pair, we used a large raw corpus of written Czech, the SYN v4 \citep{synv4,hnatkovasyn}.\footnote{\url{http://hdl.handle.net/11234/1-1846}} Specifically, we used the first 2.753.494 sentences of the corpus, which amounts to exactly 100-th of all its sentences, as classifying the corpus in its entirety (275.349.474 sentences) is above our GPU computation means. The classification took 20 hours on a single NVIDIA A100 GPU with 4 CPU threads.

\subsection{Model}
\label{sec:model}

Classification tasks on a finalized set of target variables are usually modeled as a probability distribution over K targets (possible outcomes). However, we find ourselves in an untypical situation in which the output target set is not closed yet, which requires a different perspective. If we model the problem as multi-class probability distribution, we will face an out-of-distribution problem concerning verbs which do not belong to any of the classes. We therefore model the problem as K independent binary classifiers, one for each class, of which each predicts the probability of the input belonging to the particular class in question, much like a multi-label problem. Technically, this equals to replacing the output softmax activaction function with the sigmoid activation function and accommodating the loss function accordingly, from sparse categorical cross entropy to sparse binary (focal) cross entropy,\footnote{"Focal" stands for focal loss~\cite{focal-loss}, which addresses class imbalances in training data by encouraging learning on the sparse set of hard examples (the rare positives in our case, because only one of hundreds of classes is correct) and discouraging learning from a vast majority of easy (negative) examples.} while the weights are estimated jointly by fine-tuning one shared large language model.

\subsection{Training}

Our classifier is a fine-tuned RemBERT \citep{rembert}, a rebalanced 559M-parameter mBERT,\footnote{Although BERT (110M parameters) and RemBERT ($\sim$0.5B parameters) are technically considered large language models (LLMs), they certainly rank among the modest language models w.r.t. number of parameters. Quite precisely, they belong to the masked language models (MLMs) family. Our method can however be used with any fine-tuned LLM.} with sigmoid activation function on the output layer and sparse binary focal cross entropy ($\gamma = 2.0$) to model the target class probabilities independently (see also previous Section~\ref{sec:model}). We trained our model using the Adam optimizer \cite{kingma-and-ba-2015} with defaults $\beta$'s and with a batch size of 10. The model was fine-tuned on a single NVIDA A100 GPU, using linear warm-up in the first training epoch ($6.66\%$ training steps) from 0 to peak learning rate $1\cdot10^{-5}$ and then decaying with a cosine decay schedule \cite{cosine-decay}. The model was trained for 15 epochs and we used dropout with probability $0.5$. The hyperparameters were tuned on the development set; the model achieved development set accuracy $78.67\%$ and test set accuracy $79.17\%$.

\subsection{Related Work}

We are not aware of a similar work using LLMs to classify words (and specifically, verbs) into synonym classes to enrich an existing ontology or lexicon. There are works building such resources from scratch, starting from \cite{brown-etal-1992-class} the model and its statistical,  unsupervised class hierarchy building algorithm.

The ASFALDA project (``Analyzing Semantics with Frames: Annotation, Lexicon, Discourse and Automation'')\footnote{\url{https://anr.fr/Project-ANR-12-CORD-0023}} aims at projecting English FrameNet frames to French also using machine learning but it is a recently started project and there are no published results yet.

The Predicate Matrix project \cite{lopez-de-lacalle-etal-2016-multilingual} aims at creating a resource similar to SynSemClass, by using similar resources that SynSemClass links to. The entries created automatically are not manually checked (for the most part) and we are not aware of publications describing if there were specific experiments on the comparison of the automatically created entries vs. human annotation.

There is also work on using DNNs (LSTMs specifically) to model lexical ambiguity \cite{aina-etal-2019-putting}, which is relevant for our task, but the method is not related to another existing ontological or lexical resource for training and/or fine/tuning the ML part of the system.

\begin{figure*}[!htpb]
\begin{center}
\includegraphics[width=0.999\textwidth]{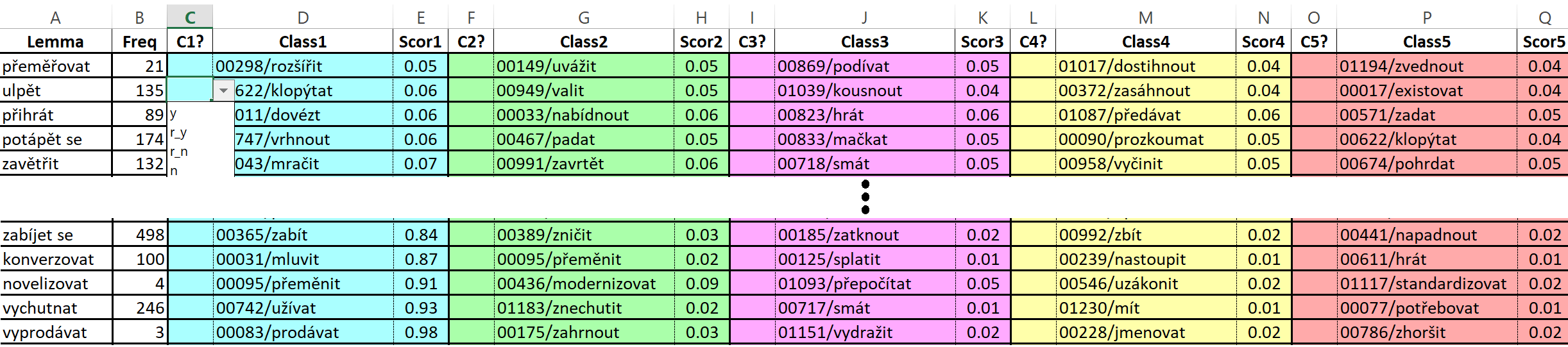}
\caption{Preprocessed data with 5 suggested classes per lemma, first and last five lines, as presented to the annotators in an Excel spreadsheet (the version with scores shown; cursor (in column C, line 3, 2$^{nd}$ data line) shows the annotation choices)}
\label{fig:dataannot}
\end{center}
\end{figure*}

\section{Post-processing with Manual Annotations}
\label{sec:manual-annotation}

%\XXX{JH EF ZU see also below}

\subsection{Input Data Preparation}
\label{indata}

In the output of the automatic classifier, each lemma has been associated with ten highest-scoring classes in which the lemma can potentially be inserted as a class member. The score is thus assigned to each lemma-class pair. These scores are numbers between 0 and 1, but it is not a probability but really just a ``score'' or a ``weight.'' The smaller the score, the less is the used LLM sure that the verb lemma belongs to the class, and vice versa - the higher the score, the more convinced it is to be added to the class.

The data as received from the classifier (3073 lemmas, with 10 class suggestions and scores for each of them) have been converted to an Excel spreadsheet to be presented to the annotators as follows:

\begin{enumerate}
    \item For each lemma (line in the resulting file), the first five classes suggested by the classifier with the highest scores as assigned by the classifier have been kept;
    \item two disjunct sets of lemmas and their class membership suggestions, with 100 lemmas each, have been randomly selected from the 3073 lemmas scored by the classifier;
    \item the two sets (called Set1 and Set2) have been converted to an Excel spreadsheet, keeping frequency information for the lemma, the five highest-scoring class membership suggestions, and the associated scores with each class;
    \item in front of each class suggestion, an extra column has been inserted with the four-way list of decisions the annotators will have to make;
    \item colors have been used to group all the information pertaining to one lemma-class pair and the decision requested;
    \item for each class suggested, a web link has been inserted in its spreadsheet cell, to allow the annotator to get to the class definition and contents (which is available on the web as shown in Fig.~\ref{slovnik2}) by a single click.
\end{enumerate}

Then, each set has been duplicated and in the copy, the scores have been deleted. The four files have then been renamed to contain the annotator abbreviation and the order number (1 for the version without scores, 2 for the version with scores (see Fig.~\ref{fig:dataannot}), i.e., in a cross-named way for the Set1 and Set2; see Table~\ref{tab:sets}).

\begin{table}[!ht]
\centering
\small
\begin{tabular}{r|c|c}
      Annotator: & A1 & A2 \\ \hline
      1$^{st}$ batch (no scores shown) & Set1 & Set2 \\ \hline
      2$^{nd}$ batch (scores shown) & Set2 & Set1 \\
      
 \end{tabular}
 \caption{Order and Assignment of Data to Annotators}
  \label{tab:sets}
\end{table}

%For the experiment, there were used Excel spreadsheets including two sets (Set1 and Set2) of automatically prepared annotation tasks which were processed by two annotators. Each set contained 100 different Czech verb lemmas. For each lemma, five automatically produced suggested classes were proposed, where perhaps the verb could belong. 

%In addition to all the suggested classes that were also in the first set, the second set (Set2) contained a clue: \XXX{JH}

%\XXX{DONE JH add example Excel lines, and possibly add a table who got what and in which order)}

\subsection{Experiment Design}
\label{design}

The Excel spreadsheets as described in the previous section (Sect.~\ref{indata}) have been sent to two annotators in two batches: first, both received the file with five suggestions for each lemma, but no scores. Each thus had 500 decisions to make (100 lemmas $\times$ 5 classifier suggestions per lemma) on a four-point scale, 0-3, denoting how strongly they recommend to include the lemma in the suggested class. The ``no'' decision corresponds to 0, ``rather\_no'' to 1, ``rather\_yes'' to 2, and ``yes'' to 3. These responses have been provided in the Excel spreadsheet as a fixed list, in order to avoid typos. In the second batch, the annotators received the other 100 lemmas, this time with scores denoting the classifier's view on the strength of the class membership recommendation, for the five classes presented.

In total, there were thus 200 lemmas manually classified twice (by the two annotators), with the classifier scores shown only for half of them to each annotator. No annotator annotated any lemma twice, and they worked independently without consulting each other. The annotators, native speakers of Czech, have been previously trained on the same task (with data coming from a different preprocessing method), so no additional training has been performed. Their pay has been based on hours worked, approx. \${8}/hour amounting to about 170\%{} of the legal minimal salary valid in 2023 in the Czech Republic.

The order and cross-assignment of the data to the annotators allowed us to measure interannotator agreement and the correlation between the annotators decisions (averaged) and the automatic classifier recommendations. Also, we could compare the speed of annotation with and without the additional clue, namely, the scores suggested by the automatic classifier. 

\begin{figure*}[!htpb]
\begin{center}
\includegraphics[width=0.95\textwidth]{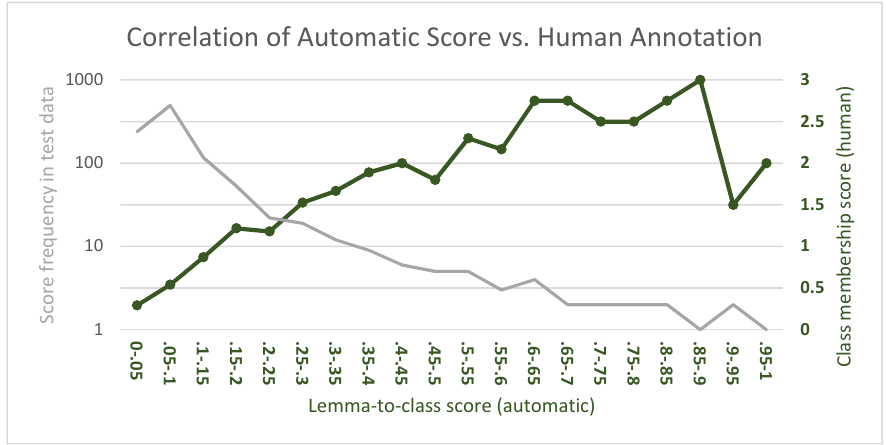}
\caption{Correlation between the automatic scores assigned to the lemma-class pairs and annotation decisions; human scores correspond to the annotation scale (3 - yes, 2 - rather\_yes, 1 - rather\_no and 0 - no) and automatic scores are bucketed (interval size: 0.05) and annotation decisions averaged in each bucket. The grey line shows the size of each bucket on a logarithmic scale.}
\label{fig:quantized-correlation}
\end{center}
\end{figure*}

\section{Results}
\label{sec:results}

%\XXX{EF: Basic statistics, the Inter Annotator Agreement (IAA)}

This section describes the results obtained as described in Sect.~\ref{sec:generating} and Sect.~\ref{sec:manual-annotation}. For the discussion of the various outputs, see Sect.~\ref{sec:discussion}.

\subsection{Human Annotation Statistics and IAA}
\label{sec:iaa}

There were 1000 pairs of Czech verb and suggested class in two sets (Set1 and Set2, see Sect.~\ref{indata}). The two annotators, A1 and A2, had to decide whether the verb could be a member of the class. Annotators could set 4 values: ``yes,'' ``rather\_yes,'' ``rather\_no'' or ``no,''. Agreement was calculated for only two values, Y and N, to which the four detailed levels of annotation have been mapped in a natural way (specifically, ``rather\_no'' has been mapped to ``N'' and ``rather\_yes'' to ``Y''). The (dis)agreement figures have been counted based on each individual decision as made by the two annotators. The resulting counts are shown in the Tab.~\ref{tab1} and agreement rate and Cohen’s $\kappa$ value in the Tab.~\ref{tab2}.

\begin{table}[!ht]
\centering
{\small
\begin{tabular}{l|c|c|c}
%\multicolumn{4}{c}{\hspace*{1.5in} } \\
\hspace*{-1.5mm} \vphantom{\Large M}A1\textbackslash{}A2 \hspace*{-1.5mm} & Y & N  & Total \\
\hline
\hspace*{-1.5mm} \vphantom{\Large M}Y & 129=66+63 & 43=15+28 & 172=81+91\\
\hline
\hspace*{-1.5mm} \vphantom{\Large M}N & 122=57+65 & 706=362+444 \hspace*{-1.5mm} & 828=419+409\\
\hline
\hspace*{-1.5mm} \vphantom{\Large M}Total & \hspace*{-1.5mm} 251=123+128 \hspace*{-1.5mm} &  \hspace*{-1.5mm} 749=377+372 \hspace*{-2.5mm} & 1000=500+500 \hspace*{-1.5mm}  \\
  \end{tabular}
}
\caption{Annotation statistics: counts shown for the 1000 annotation decisions (500 from Set1, 500 from Set2). Mappings used: y $\rightarrow$ Y, r\_y $\rightarrow$ Y, r\_n $\rightarrow$ N, n $\rightarrow$ N. Counts are presented in the cells as Total-$xy$=Set1-$xy$+Set2-$xy$, where $x, y \in \{$Y, N$\}$.}
\label{tab1}
\end{table}

\begin{table}[ht]
\centering
{\small
\begin{tabular}{l|c|c}
\vphantom{\Large M} & IAA  & Cohen's $\kappa$ \\
 \hline
\vphantom{\Large M}All & 0.83 & 0.51\\
\hline
\vphantom{\Large M}Set1 & 0.86 & 0.56 \\
\hline
\vphantom{\Large M}Set2 & 0.81 & 0.46 \\
 \end{tabular}
}
\caption{Inter-annotator agreement and Cohen's $\kappa$ between annotators, for the 500 decisions each annotated by both annotators, with the scaled values mapped to Y/N only.}
\label{tab2}
\end{table}

\begin{table}[!ht]
\centering
{\small
\begin{tabular}{l|c|c}
%\multicolumn{2}{c}{\hspace*{1.5in} } \\
 & Batch 1 (no scores) & Batch 2 (with scores) \\
 \hline
\vphantom{\Large M}A1 & 192 & 174\\
 \hline
\vphantom{\Large M}A2 & 210 & 210 \\
 \hline
\vphantom{\Large M}Average & 201 & 192 \\
 \end{tabular}
}
\caption{Time of annotation by annotators A1 and A2, in minutes. Batch 1 is Set1 and Set2 without showing the scores assigned by the automatic classifier, Batch 2 shows the scores.}
\label{tabtime}
\end{table}

\subsection{Human Annotation Time}
\label{sec:humantime}

The annotators have been asked to record the time it took them to annotate the data. Each Set entailed 500 decisions, which took slightly over three hours on average. The detailed breakdown is shown in Tab.~\ref{tabtime}.

%\begin{table}[ht]
%\centering
%{\small
%\begin{tabular}{l|c|c}
%\multicolumn{3}{c}{\hspace*{0.9 in} } \\
%  & \exfunc{A1} & \exfunc{A2}\\
%Order1 &  \exfunc{Set1-without scores} & %\exfunc{Set2-without scores} \\
%Order2   & \exfunc{Set2-with scores} & %\exfunc{Set1-with scores} \\
% \end{tabular}
%}
%\caption{statistiky }
%\label{tab2}
%\end{table}

\subsection{Correlation between the Scores of the Automatic Classifier and the Human Annotation}
\label{sec:correlation}

To find if there is a relationship between the automatic scores and manually annotated data, we used the Pearson’s correlation (Pearson’s r) coefficient. Automatic scores and human annotations were found to be moderately correlated ($r(998) = .44$, $p < .001$). A Spearman's correlation was also run to determine the relationship between $1000$ automatic scores and human annotations. There was weak to moderate monotonic correlation between automatic scores and human annotations ($\rho = .39$, $n = 1000$, $p < .001$).

We visualize the correlation between the automatic scores assigned to the lemma-class pairs and annotation decisions in Fig.~\ref{fig:quantized-correlation}; human scores correspond to the annotation scale (3 - yes, 2 - rather\_yes, 1 - rather\_no and 0 - no) and automatic scores are bucketed (interval size: 0.05) and annotation decisions averaged in each bucket, effectively smoothing out the curve by reducing variance. The Pearson correlation between scores and human annotations averaged per bucket is $r(18) = .79$ ($p < .001$) and Spearman's ranked correlation is $\rho = .76$ ($n = 20$, $p < .001$).

\section{Discussion of Results}
\label{sec:discussion}

It is well known that trained annotators often create high-quality data, needed for many NLP applications, although their services are generally expensive.The experiment described here was designed to answer several questions:

\begin{itemize}
    \item What is the usual inter-annotator agreement for the human assignment of verbs to classes, using pre-annotated data?
    \item Can a heuristics be defined to indicate which pre-assigned lemma-class pairs the annotators can trust and to what extent?
    \item Does the scoring mechanism, which provides scores for each of the lemma-class relation strength, make the annotation more efficient? 
    \item Is the automatic classifier for computing the relation strength between an unknown lemma and an existing class(es), as described in Sect.~\ref{sec:generating}, in any way correlated with the human decisions made by experienced annotators?
\end{itemize}

As seen from Sect.~\ref{sec:iaa} (Tab.~\ref{tab2}), the inter-annotator agreement is relatively high (0.83 on average over the two Sets), but the Cohen's kappa $\kappa$ is low (0.51 on average over the same two Sets annotated). However, the low kappa is caused by the highly skewed distribution of the decisions,\footnote{Almost 4:1 - the average number of (mapped) ``No'' decisions is 788,5 out of 1000.} the most of which lead to the rejection of the assignment of the lemma to the suggested class, caused mainly by the selection of a fixed number of five suggestions per lemma regardless of the score computed by the classifier. It would be possible - by using more pairs of annotators - to optimally select the number of suggested classes (i.e., most probably between 1 and 5), but it would only be relevant for the current number of classes in the ontology. As the ontology grows, the number of rejections will be different and the optimal number of classes might change.

%For any heuristics in terms of a threshold which would signal a good candidate vs. a rejection of all classes, the same consideration as for the number of classes to present to the annotators holds.

For the size of the ontology on which it has been tested, the threshold separating the Yes/No decision (with the highest uncertainty being around the average of 0-3, i.e., 1.5) seems to be around 0.3 (see Fig.~\ref{fig:quantized-correlation}). However, due to the linearity of the correlation (which by itself is a positive result for the classifier--see below), it would still be necessary to provide careful manual inspection results on both sides of the threshold. The same holds for setting any thresholds at the low or high ends of the classifier score scale.
%podivej se prosimte na posledni odstavec casti Limitations, tam je to s tou coverage cz a eng nejak zamotany OK podivam, jinak jsem ty tvoje body uz prosel cely za chvili poslu email. Ok, 
In terms of annotation efficiency (providing scores to the annotators vs. not providing them), the result is largely negative. A small speedup has been observed only for A1, with A2 consuming the same time for both Sets. The absolute time as recorded by the annotators per lemma (i.e., 5 times the single decisions time, which was $366 \times 60 \div 1000 \approx 22$ sec. for A1 and $420 \time 60 \div 1000 \approx 25$ sec. for A2) is about two minutes. This is in fact a positive finding which means that the whole set of pre-classified lemmas, as processed by the classifier (3073 lemmas) would be finished within approx. 6000 minutes (100 hours) per annotator, i.e., within 200 hours with double annotation, plus adjudication time.

Finally, the correlation between the automatic classifier and the human annotation is very strong. Of course, the bucketing to the 0.05 interval improves the correlation (see Sect.~\ref{sec:correlation}), but in any case, it seems that the classifier
% JS: Jelikož si Reviewer #3 stěžoval, že to není žádné "large", tak jsem tuhle vsuvku dala pryč. Asi by se naopak dalo oponovat, že to je vlastně super, že už jen při použití "malého" LM už máme nějaké výsledky, ale raději bych do této nové diskuze nezabředávala.
%, using a relatively large LM,
% JS: V následující větě correlation of the unknown lemma -> affiliation of the unknown lemma
% Byl to asi spíš omyl, protože to jednak není korelace (je to nějaké skóre) a jednak se nám slovo korelace opakuje.
is able to assign the score denoting the strength of affiliation of the unknown lemma to a class with high correlation to the human annotation decisions.

\section{Conclusions and Future Work}
\label{sec:conclusions}

As discussed in the previous section (Sect.~\ref{sec:discussion}), the strongest result achieved in this study is the correlation between the classifier scoring buckets and the human decisions (Fig.~\ref{fig:quantized-correlation}, Sect.~\ref{sec:correlation}). While the scores themselves, when presented to the annotators, do not seem to bring higher efficiency, the selection of the classes and their presentation to the annotators (Sect.~\ref{sec:generating}, Sect.~\ref{indata}) result in a reasonable time for the annotation of several thousand previously unseen (unassigned lemmas) to the ontology. Finally, there is no strong heuristics (for the score thresholds) that would allow to assign any unseen words to existing classes automatically -- a human post-inspection and annotation is needed across the whole (or almost whole) range of scores as produced by the classifier, given the linear correlation.

In the future, we plan to repeat the experiment for a larger ontology (i.e., test the effort needed for sustainable development and maintenance for such an event-type ontology when it reaches high coverage), possibly with larger LMs or with some additional fine tuning given the large(r) coverage at such future time.

%\section{Appendices}

%\XXX{LEFT ONLY FOR INFO - perhaps put the tables and output form the classifier as appendix (and then to LINDAT if accepted)?}

%\XXX{Use \textbackslash{}appendix before any appendix section to switch the section numbering over to letters. See Appendix~\ref{sec:appendix} for an example.
%}

\section*{Limitations}

%JS Já si vlastně nejsem jistá, tak první odstavec dám pryč (promiň Zdeni, že Tvoje reference přijde vniveč) a možná to vrátím, až se ujistím. Ono to vlastně není až tak Limitation, jako spíš něco do Related Work. Ale jak říkám, nejsem si jistá, jak je to vlastně se stavem korpusů v UD (ověřuji).

%za mne by se to musela preformulovat, podle mne i v UD je vetsina korpus aspon doanotovana rucen, pokud neni rucne komplet (jako je cestina).

%While it is not uncommon to use machine learning for automatic annotation of linguistic data (e.g., many corpora in UD has been automatically tagged on the morphological level in the Universal Dependencies project \citep{NideUniversalDependencies2016, BoSeOverviewIWPT2020}, it is still quite rare to use machine learning as an aid for annotation process.

We advocate for a moderate and restrained usage of automatic guiding methods and we must advise caution to take the automatic output with a grain of salt, both qualitatively and quantitatively. First, the classifier predictions can fall far from gold labels and should not be considered as such. Second, although measures have been taken to mitigate the out-of-distribution classification problem, one should be aware of the fact that by the very nature of the problem, which is annotation of completely new, possibly out-of-distribution data, the classification predictions are not to be trusted indiscriminately and should subsequently be approved by the annotators. The annotators should be instructed to consider the suggestions as election votes. Furthermore, we should refrain from overly automating the entire annotation process so as to achieve high alignment with the machine learning suggestions, which might lead to trivial and unimaginative annotations from the linguistic perspective. Finally, exhausting the informativeness of the pre-trained (albeit fine-tuned) model might prevent further learning from the annotated data.

Another limitation of the results, or the interpretation of the results, is the fact that the model is trained on an actual state of the ontology. It means that in fact the classifier would have to be retrained after adding a single new class or even a new lemma to an existing class; while in practice it would be OK to process several lemmas at once, it is still a limitation given the non-negligible training and prediction time (20 hours on a single GPU) which cannot be parallelized (see Sect.~\ref{sec:generating}).
% Přidala "and prediction", protože trénování je pravda sice několik hodin, ale ten údaj 20 hodin se vztahuje k následným predikcím pro nezpracovaná lemmata na velkém raw corpusu.

In addition, the correlation might decrease and the thresholds shift as the size (and thus coverage) by the ontology grows, since the unseen lemmas will be increasingly rare, with possibly less data available in the LM to reliably estimate the scores. Conversely, for ontologies with much smaller coverage (e.g., for ontologies the development of which has just started) the same shifts in correlation and thresholds are likely.

Finally, the whole experiment has been performed on Czech due to the lower coverage of the ontology than for English, and also in order to explore a morphologically rich language with a high form-to-lemma ratio. Results for other languages might differ.

%\XXX{ACL 2023 requires all submissions to have a section titled ``Limitations'', for discussing the limitations of the paper as a complement to the discussion of strengths in the main text. This section should occur after the conclusion, but before the references. It will not count towards the page limit.
%The discussion of limitations is mandatory. Papers without a limitation section will be desk-rejected without review.
%While we are open to different types of limitations, just mentioning that a set of results have been shown for English only probably does not reflect what we expect. 
%Mentioning that the method works mostly for languages with limited morphology, like English, is a much better alternative.
%In addition, limitations such as low scalability to long text, the requirement of large GPU resources, or other things that inspire crucial further investigation are welcome.}

\section*{Ethics Statement}

The human subjects used in this study have been experienced, trained annotators who have been in personal contact with the authors, and who have been recruited by a call specifically suited for the experiment and study presented here. The call has been sent to all trained annotators already working with the authors, and volunteers have been asked to respond, on a first-come first-chosen basis. The pay has corresponded to the standard pay for similar annotation tasks taking also the relatively short notice into consideration (for the numbers, see Sect.~\ref{design}). Both annotators were males; this is a possible shortcoming, but there were no female volunteers and from the previous cooperation (with a mixed team of female and male annotators), no differences in the annotation results have been observed.

No personal information has been among the lemmas extracted and used for the preselection. The data for the LLM might have contained it, but it would not show because the experiment and the ontology is currently limited to common verbs which do not describe any personal names or other personal information.

%OK snad vtom jednovetnem popisku neudelam nejakou chybu.. jdu na to..ok

%\XXX{JH response to ethics guidelines, pay to annotators, annotator biases, no harm...}

%\XXX{Scientific work published at ACL 2023 must comply with the ACL Ethics Policy.\footnote{\url{https://www.aclweb.org/portal/content/acl-code-ethics}} We encourage all authors to include an explicit ethics statement on the broader impact of the work, or other ethical considerations after the conclusion but before the references. The ethics statement will not count toward the page limit (8 pages for long, 4 pages for short papers).}

\section*{Acknowledgements}

The work described herein has been supported by the Grant Agency of the Czech Republic under the EXPRO program as project “LUSyD” (project No. GX20-16819X) and uses resources hosted by the LINDAT/CLARIAH-CZ Research Infrastructure (projects LM2018101 and LM2023062, supported by the Ministry of Education, Youth and Sports of the Czech Republic).

We would like to thank the annotators Petr Kujal and Tomáš Razím for their valuable work and invaluable input, as well as the three anonymous reviewers for their insightful comments.

\bibliography{anthology,custom}
\bibliographystyle{acl_natbib}

\onecolumn

\appendix
\section*{Appendices}

\section*{Classifier and Annotation Results}
\label{sec:appendix}

We are providing Supplemental material with the raw classifier file and the human annotation results. The open-source code and the data itself are provided at GitHub  (\url{https://github.com/strakova/synsemclass_ml}). Here, technical description of the supplemental material is provided on top of what has been mentioned in the paper.

\subsection*{Classifier output}

The raw output of the classifier, with the 3073 previously unseen (unassigned) lemmas and their classification scores to 10 closest classes, is attached in the Supplemental material file (file \texttt{all\_buckets\_2753494.txt}).

The file contents is structured as follows (each lemma and classifier scores are on a single line):

\vspace*{3mm}
\noindent
{\tt \footnotesize
lemma freq-in-data max-score suggested-class-1 score-class-1 ... suggested-class-10 score-class-10 
}

\vspace*{3mm}

where 

\vspace*{3mm}

\noindent \texttt{lemma} 

is the lemma which has been classified to all the available classes in SynSemClass

\noindent \texttt{freq-in-data} 

is the frequency of the lemma in the dataset used for building the LM

\noindent \texttt{max-score} 

is the maximum score (score of the first class in the list)

\noindent \texttt{suggested-class-n} 

is the ID and name (\& Czech sense ID) of the n-th best class assigned to the \texttt{lemma} by the classifier 

\noindent \texttt{score-class-n} 

is the score assigned to the (\texttt{lemma,suggested-class-n}) pair.

\subsection*{Annotation Results}

The annotation results are presented as four Excel Spreadsheets, named \texttt{law-Am-n.xlsx}, where \texttt{m} is the annotator ID and \texttt{n} is the batch number (i.e., the lemmas and classes are identical for \texttt{A1-1} and \texttt{A2-2} and for \texttt{A1-2} and \texttt{A2-1}, except for the presence of scores and differing also of course in the assigned \texttt{y/r\_y/r\_n/n} labels by the annotators). 

Each Excel file has 100 content lines (100 lemmas and 5 best classes for each as classified by the pre-annotation tool):

\vspace*{3mm}

\noindent \texttt{Lemma} 

is the lemma being classified

\noindent \texttt{Freq} 

is the (informative-only) frequency of the lemma in the training text

\noindent \texttt{Cn?}

is the column where the annotators recorded their decisions

\noindent \texttt{Classn}

is the ID of the class (clickable)

\noindent \texttt{Scorn}

is the score of the lemma-class affiliation by the classifier (in ...-2.xlsx files only)

\noindent \texttt{AnnotatorComment}

is an optional annotator's comment.

%\XXX{Original data output here? (3000 - too long, maybe as supplemental ZIP file with submission is better. Or Code structure (at least)? Hyperparameter table?  Or to text?
%What else to put here?}

\end{document}